\documentclass{article}
\usepackage{spconf,amsmath,graphicx}
\usepackage{amsmath,amssymb,amsfonts}
\usepackage{graphicx}
\usepackage{textcomp}
\usepackage{subfig}
\usepackage{xcolor}
\usepackage{algpseudocode}
\usepackage{algorithm}
\usepackage{makecell}
\usepackage{hyperref} 
\usepackage[export]{adjustbox}

\title{Archetypal Analysis for Binary Data}
\name{A. Emilie J. Wedenborg  and Morten Mørup\thanks{This work was supported by the Danish Data Science Academy, which is funded by the Novo Nordisk Foundation (NNF21SA0069429) and VILLUM FONDEN (40516).}}
\address{The Technical University of Denmark\\ DTU Compute - Cognitive Systems Section\\ \texttt{\{aejew,mmor\}@dtu.dk}}
\makeatletter

\def\ps@IEEEtitlepagestyle{
  \def\@oddfoot{\mycopyrightnotice}
  \def\@evenfoot{}
}
\def\mycopyrightnotice{
  {\footnotesize xxx-x-xxxx-xxxx-x/xx/\$31.00~\copyright~2018 IEEE\hfill} 
  \gdef\mycopyrightnotice{}
}

\@ifundefined{showcaptionsetup}{}{
 \PassOptionsToPackage{caption=false}{subfig}}
\usepackage{subfig}
\makeatother

\usepackage{eso-pic}
\newcommand\AtPageUpperMyright[1]{\AtPageUpperLeft{
 \put(\LenToUnit{0.085\paperwidth},\LenToUnit{-2cm}){
     \parbox{\textwidth}{\raggedright\fontsize{9}{11}\selectfont #1}}
 }}
\newcommand{\conf}[1]{
\AddToShipoutPictureBG*{
\AtPageUpperMyright{#1}
}
}

\conf{Copyright 2025 IEEE. Published in ICASSP 2025 – 2025 IEEE International Conference on Acoustics, Speech and Signal Processing (ICASSP), scheduled for 6-11 April 2025 in Hyderabad, India. Personal use of this material is permitted. However, permission to reprint/republish this material for advertising or promotional purposes or for creating new collective works for resale or redistribution to servers or lists, or to reuse any copyrighted component of this work in other works, must be obtained from the IEEE. Contact: Manager, Copyrights and Permissions / IEEE Service Center / 445 Hoes Lane / P.O. Box 1331 / Piscataway, NJ 08855-1331, USA. Telephone: + Intl. 908-562-3966.} 

\begin{document}
\ninept
\maketitle
\begin{abstract}
Archetypal analysis (AA) is a matrix decomposition method that identifies distinct patterns using convex combinations of the data points denoted archetypes with each data point in turn reconstructed as convex combinations of the archetypes. AA thereby forms a polytope representing trade-offs of the distinct aspects in the data. Most existing methods for AA are designed for continuous data and do not exploit the structure of the data distribution. In this paper, we propose two new optimization frameworks for archetypal analysis for binary data. i)  A second order approximation of the AA likelihood based on the Bernoulli distribution with efficient closed-form updates using an active set procedure for learning the convex combinations defining the archetypes, and a sequential minimal optimization strategy for learning the observation specific reconstructions. ii) A Bernoulli likelihood based version of the principal convex hull analysis (PCHA) algorithm originally developed for least squares optimization. We compare these approaches with the only existing binary AA procedure relying on multiplicative updates and demonstrate their superiority on both synthetic and real binary data. Notably, the proposed optimization frameworks for AA can easily be extended to other data distributions providing generic efficient optimization frameworks for AA based on tailored likelihood functions reflecting the underlying data distribution.
\end{abstract}
\section{Introduction}
\label{sec:intro}
Archetypal Analysis (AA) \cite{Cutler1993ARCHETYPALANALYSIS} is a matrix decomposition technique known for identifying unique characteristics called archetypes. These archetypes, which represent the vertices of an optimally learned polytope, are constrained to lie within the convex hull of the data, forming what is known as the principal convex hull \cite{Mrup2012ArchetypalMining}. This distinctive feature of AA enables it to capture the most prominent aspects of the data, offering an easily interpretable model where each data point is represented as a combination of trade-offs between the archetypes. 
Despite its potential, finding the optimal archetypes is a challenging problem that involves solving a non-convex optimization problem. A solution is to split the problem into smaller convex quadratic programming (QP) subproblems that are iteratively alternatingly solved to converge to a (local) minima. Some of the attempts to optimize this procedure include fast gradient projection methods based on the principal convex hull analysis (PCHA)\cite{Mrup2012ArchetypalMining}, optimized loss functions, where a Huber loss function is used together with an iterative, reweighed least squares strategy \cite{Chen2014FastLearning}, faster QP-solvers \cite{Chen2014FastLearning}, decoupling of data and archetypes, enforcing unit normalization and learning the convex hull inside the unit sphere \cite{Jieru2018OnlineAnalysis} and data driven selection and processing steps for faster convergence \cite{Han2022ProbabilisticAnalysis,Mrup2012ArchetypalMining,Mair2019CoresetsAnalysis,Bauckhage2009MakingPractical}. These extensions have been tailored for continuous data sources based on least squares minimization.
Discrete data types have not been explored as extensively in Archetypal Analysis.
While previous efforts have aimed at expressing the model in probabilistic terms \cite{Seth2016ProbabilisticAnalysis}, with extensions to discrete distributions, these endeavors have relied on multiplicative updates that are known to converge slowly \cite{Gillis2008NonnegativeProblem,Nielsen2014Non-negativeBrain}, involved transforming the data \cite{Gimbernat-Mayol2021ArchetypalGenetics}, or used AA with archetypes constrained to actual cases \cite{Cabero2020}.
In this paper, we propose two efficient optimization frameworks for archetypal analysis for binary data. Firstly, we propose a novel framework for AA that exploits i) the sparsity of the convex combinations learned to define the archetypes by use of an active set algorithm \cite{Bro1997AAlgorithm} tailored to convex constraints and ii) the low dimensional structure of the matrix forming their reconstruction which we demonstrate can be efficiently solved by sequential minimal optimization (SMO) originally used in the context of support vector machines \cite{Platt1998SequentialMachines}. We thereby establish efficient closed-form updates for the two convex sub-problems used alternatingly to solve for AA. We exploit how these updates can be applied to arbitrary likelihood specifications by use of second order likelihood expansions. While our focus is on Bernoulli distributed binary data, the framework can easily be expanded to any likelihood function. We identify and demonstrate the same attribute in the Principal Convex Hull Algorithm (PCHA) \cite{Mrup2012ArchetypalMining} where we derive new gradients tailored for a Bernoulli Likelihood. 
Specifically, we:
\begin{itemize}
\item Derive an efficient procedure for AA inference exploring sparsity of active set defining the convex combinations forming the archetypes and SMO updates for the low-dimensional observation specific reconstruction defining convex combinations of the archetypes.
\item Use the approach for generic AA optimization based on quadratic expansion of AA likelihoods.
\item Extend the PCHA algorithm to Bernoulli likelihood optimization enabling principled analyses using PCHA of binary data.  
\item Showcase on synthetic and real datasets the superiority of the optimization procedure when compared to the existing multiplicative updates for Bernoulli likelihood optimization.
\end{itemize}

\section{Methods}
\subsection{Likelihood Optimized Archetypal Analysis}
For a data matrix of features by observations $X \in \mathbb{R}^{M \times N}$, the primary goal of archetypal analysis (AA) is to decompose the data according to \cite{Cutler1993ARCHETYPALANALYSIS}:

\begin{equation} \label{eq:AAClassic}
\begin{array}{clll}
\min _{\mathbf{C}, \mathbf{S}} & L(\mathbf{X},\mathbf{R})&
\text { s.t. } & \mathbf{R} = \mathbf{XCS} \\
&&&
c_{j, k} \geq 0, \quad s_{k, j} \geq 0 \\
&&& \sum_j c_{j,k} = 1, \quad \sum_k s_{k,j} = 1
\end{array}
\end{equation}

such that $\mathbf{C} \in \mathbb{R}^{N\times K}$ and $\mathbf{S} \in \mathbb{R}^{K\times N}$ are matrices formed by columns constrained to reside on the standard simplex defining convex combinations.

Depending on the specification of the loss $L(\mathbf{X},\mathbf{R})$, we can express various probability distributions in context of archetypal analysis (AA) allowing to customize the model for specific data types, see also \cite{Seth2016ProbabilisticAnalysis}. We presently detail two loss function specifications encompassing the conventional least squares (eq. \ref{eq:LSloss}) and Bernoulli (cross-entropy)  loss (eq. \ref{eq:Berloss})

\begin{equation}\label{eq:LSloss}
    L =\sum\limits_{i,j} ||(x_{i,j}-r_{i,j})||_F^2,
\end{equation}

\begin{equation}\label{eq:Berloss}
 L = \sum\limits_{i,j} -x_{i,j} \ln(r_{i,j})
    -(1-x_{i,j})\ln(1-r_{i,j}).
\end{equation}
In the Bernoulli case $\mathbf{R}$ is redefined as $\mathbf{R} = \mathbf{PCS}$, where $\mathbf{P}$ is defined as $ \mathbf{P} = \mathbf{X}+\varepsilon-2\cdot \mathbf{X}\varepsilon$, for $\varepsilon = 1e^{-3}$.
Importantly, we provide a generic efficient optimization framework relying on solving the two convex sub-problems respectively optimizing for $\mathbf{S}$ and $\mathbf{C}$. 
The code is publicly available on \href{https://github.com/Wedenborg/Archetypal-Analysis-For-Binary-Data}{\texttt{github.com/Wedenborg}}.

\begin{figure*}[htbp]
\centering
\subfloat[]{\includegraphics[width=0.32\textwidth]{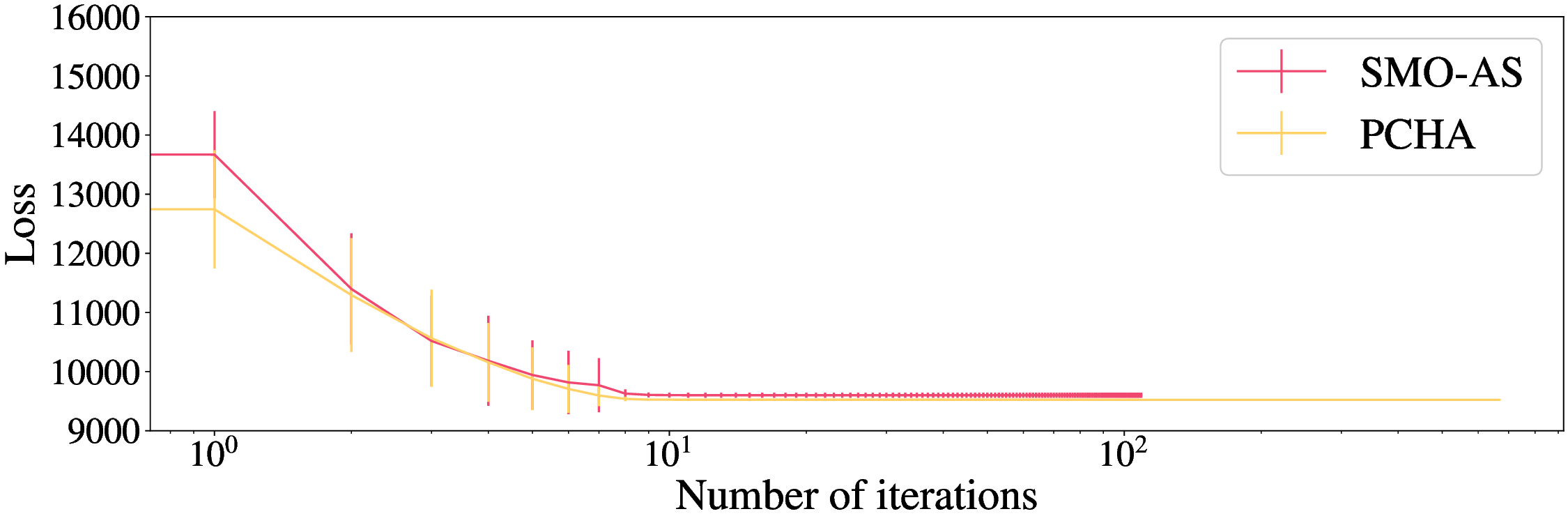}\label{fig:ConvGauSynth}}
\subfloat[]
{\includegraphics[width=0.32\textwidth]{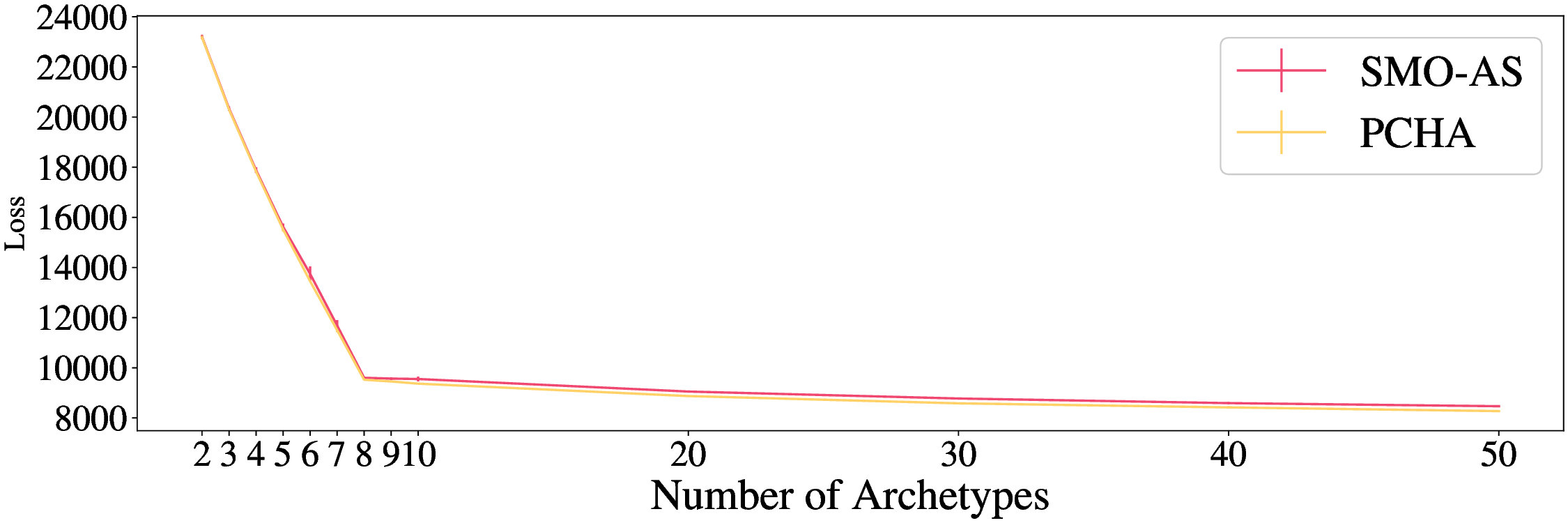}\label{fig:LossGauSynth}}
\subfloat[]{\includegraphics[width=0.32\textwidth]{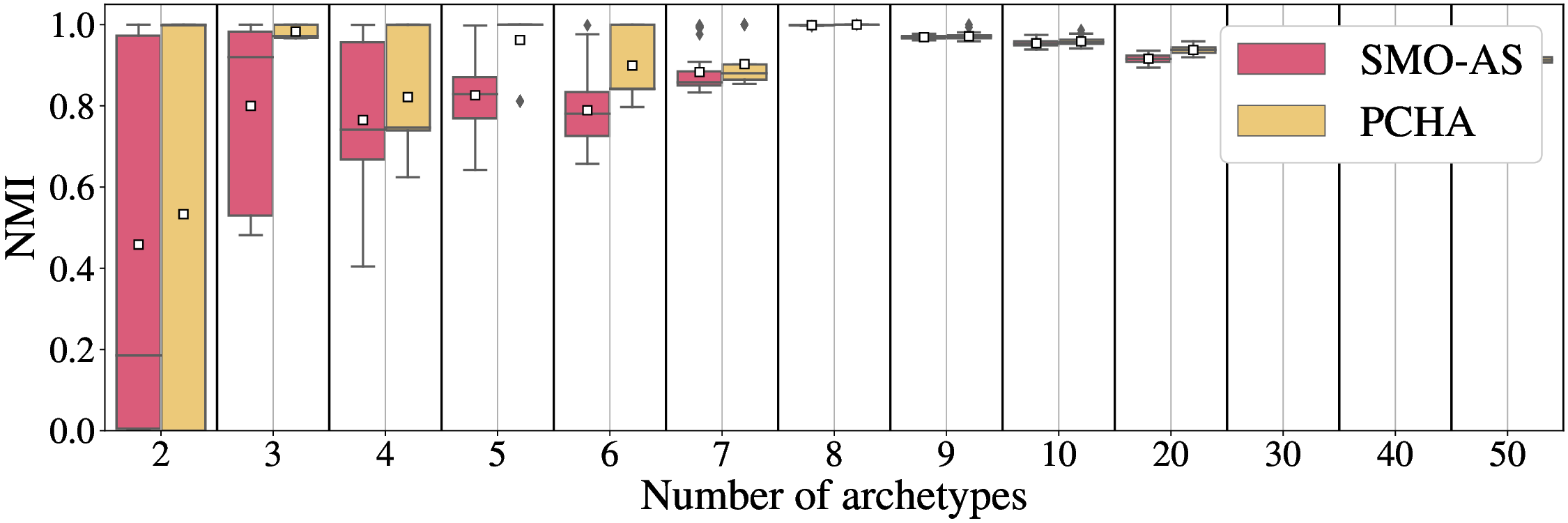}}
\vspace{-4mm}
\subfloat[]
{\includegraphics[width=0.32\textwidth]{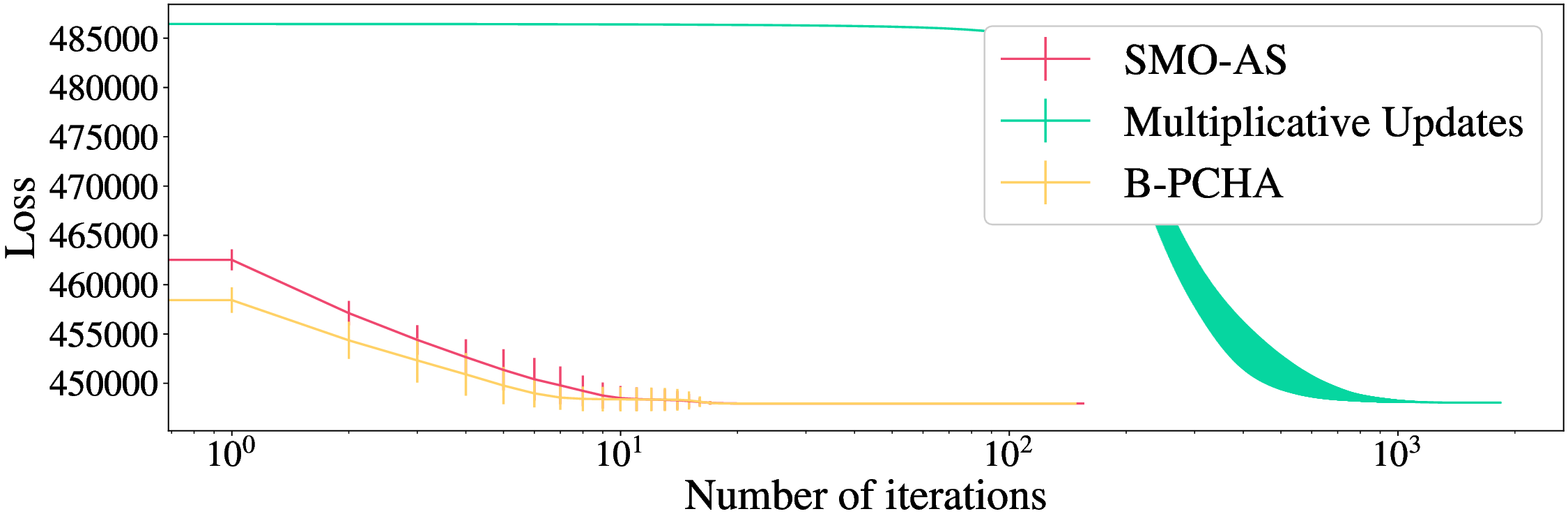}\label{fig:ConvBerSynth}}
\subfloat[]{\includegraphics[width=0.32\textwidth]{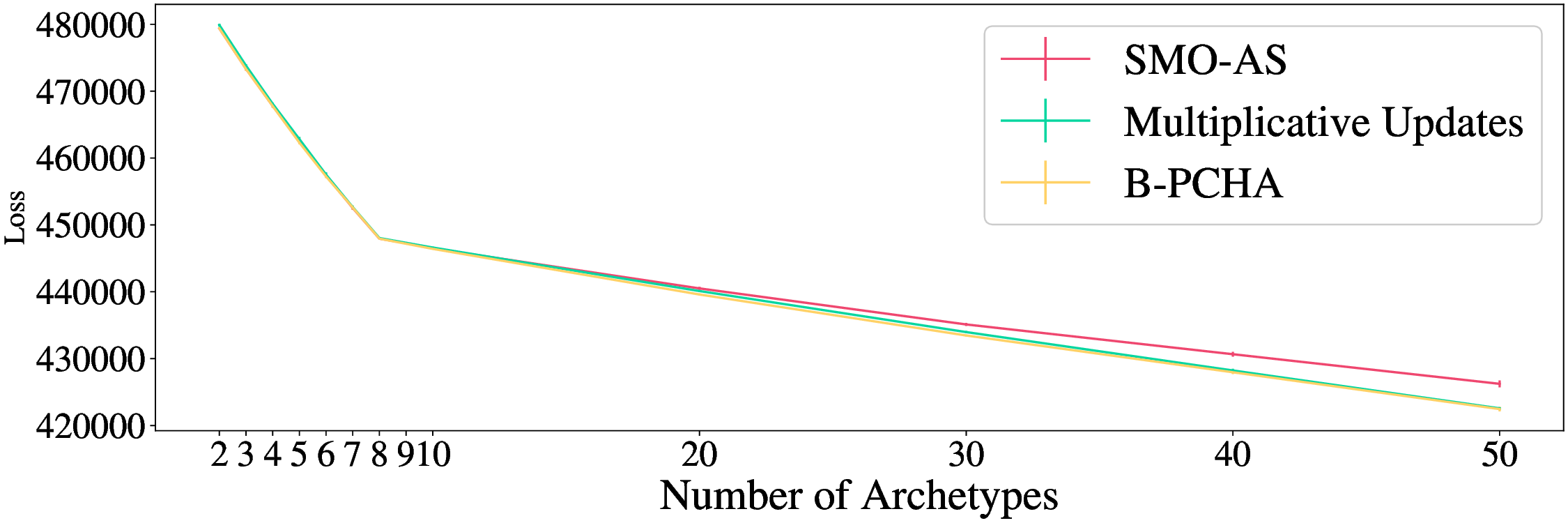}\label{fig:LossBerSynth}}
\subfloat[]{\includegraphics[width=0.32\textwidth]{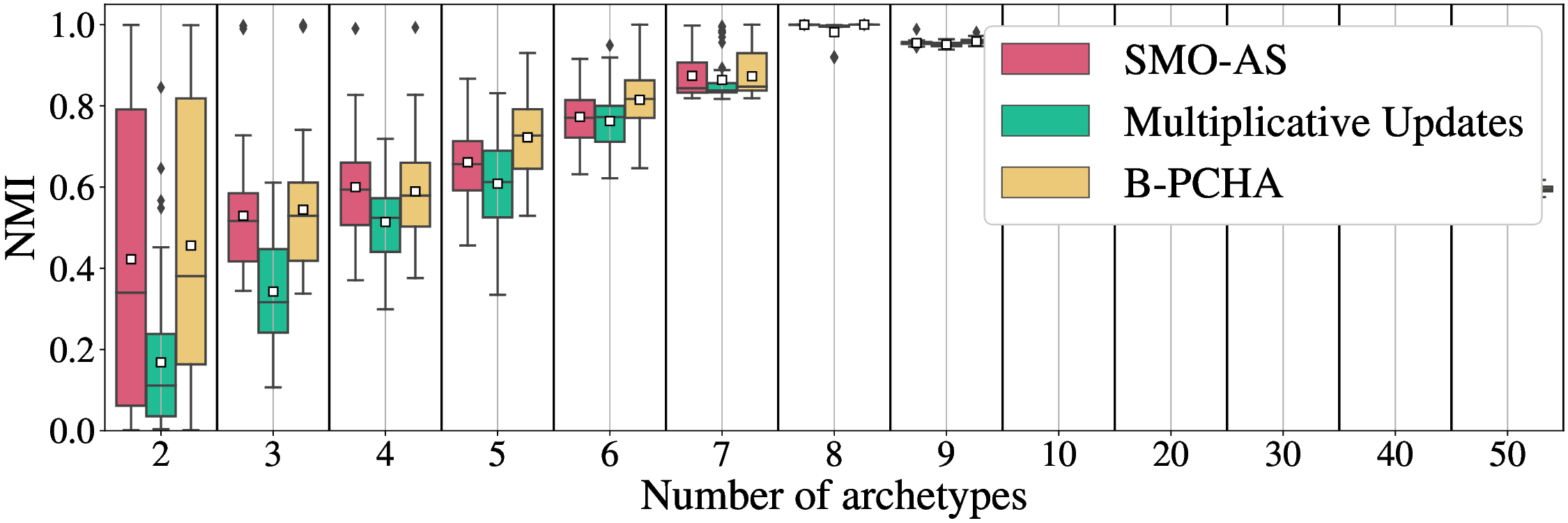}\label{fig:NMIBerSynth}}
\caption{\textbf{Top panel:} Results of the models on synthetic Gaussian data. \textbf{Bottom panel:} Results of the models on synthetic Bernoulli data. The leftmost column represents the loss convergence properties. The middle plots shows the loss for different numbers of archetypes. The left column displays the NMI which in this case is used as a measure of the stability of the solution.}\label{fig:synth}
\vspace{-2mm}
\end{figure*} 

\subsection{$\mathbf{S}$-update by Sequential Minimal Optimization}
Sequential Minimal Optimization (SMO) was first proposed in \cite{Platt1998SequentialMachines}. This method splits large Quadratic Problems (QP) into the smallest possible QP problems. The method is especially suited for updating the $\mathbf{S}$ matrix as this is typically a low-dimensional dense matrix, representing the data in terms of convex combinations of the archetypes. In eq. (\ref{eq:Fs}) - (\ref{eq:alpha}) the generic framework for an arbitrary distribution is shown. By applying a second order Taylor expansion to the likelihood functions specified in the equations \ref{eq:LSloss} and \ref{eq:Berloss}, the expression for the likelihood in terms dependent on $\mathbf{S}$ decouples into independent column specific terms $f(\mathbf{s}_j)$ given by 
\begin{equation} \label{eq:Fs}
    f(\mathbf{s}_j) \approx const. -\mathbf{d}_j^\top\mathbf{s}_j+\frac{1}{2}\mathbf{s}_j^\top\mathbf{H}^{(j)}\mathbf{s}_j
\end{equation}
$\mathbf{d}$ and $\mathbf{H}$ are derived from the second order Taylor approximation and their definitions are given in equations \ref{eq:d_LS}-\ref{eq:H_BER}. For the least squares (i.e., Gaussian likelihood with unit variance) they take the form: 
\begin{equation}\label{eq:d_LS}
\begin{aligned}
d_{k,j} &= - 2(\sum\limits_{i,m} c_{m,k} x_{i,m} x_{i,j} + \sum\limits_l   c_{l,k} x_{i,l} r_{i,l} \\
-& \sum\limits_{k^{\prime},k^{\prime\prime}} h^{(j)}_{k^{\prime},k^{\prime \prime}}s_{k,j}),
\end{aligned}
\end{equation}
\begin{equation}\label{eq:H_LS}
    h^{(j)}_{k^{\prime},k^{\prime \prime}} = \sum\limits_{l,l^{\prime},i}   c_{l,k^\prime} x_{i,l} x_{i,l^{\prime}} c_{{l^\prime},k^{\prime\prime}},
\end{equation}
whereas for the Bernoulli likelihood they become: 
\begin{equation}\label{eq:d_BER}
\begin{aligned}
d_{k,j} &= - \sum\limits_{i} ( \frac{x_{i,j}}{r_{i,j}} +\frac{(1-x_{i,j})}{(1-r_{i,j})})\sum\limits_{m}p_{i,m}c_{m,k}\\ &-\sum\limits_{k^{\prime},k^{\prime\prime}}h^{(j)}_{k\prime,k\prime\prime}s_{j,k},
\end{aligned}
\end{equation}

\begin{equation}\label{eq:H_BER}
    \begin{aligned}
       h^{(j)}_{k^{\prime},k^{\prime \prime}} =& \sum\limits_{i} (\frac{x_{i,j}}{(r_{i,j})^2}+ \frac{(1-x)_{i,j}}{(1-r_{i,j})^2})\\
     \sum\limits_{m}& p_{i,m}c_{m,k^\prime}\sum\limits_{i,m^{\prime}}p_{i,m^{\prime}}c_{m^{\prime},k^{\prime\prime}}.
    \end{aligned}
\end{equation}

By utilizing the SMO framework, we look at the archetypes in pairs and update the pairs individually. To ensure convergence all possible pairs of archetypes are considered. This means that we substitute $\mathbf{s}_j$ with $t_j=\sum_{d \in \{k^\prime, k^{\prime\prime}\}} s_{d,j}$, where $k^\prime$ and $k^{\prime\prime}$ represent any given pair of archetypes. For the SMO updates we optimize a weight parameter, $\alpha_j\in[0,1]$ such that $s_{k^{\prime},j}=t_j \alpha_j$ and $s_{k^{\prime\prime},j}=t_j (1-\alpha_j)$, considering the re-parameterized loss function 
\begin{equation} \label{eq:Fs_matrix}
    \begin{aligned}
        F(\mathbf{s}_j) \approx const - \begin{bmatrix}
            d_{k^{\prime},j} \\
            d_{k^{\prime\prime},j}
        \end{bmatrix}^\top \begin{bmatrix}
             t_j \alpha_j \\
             t_j(1-\alpha_j)
        \end{bmatrix}+\\\frac{1}{2}
        \begin{bmatrix}
             t_j\alpha_j \\
             t_j(1-\alpha_j)
        \end{bmatrix}^\top \begin{bmatrix}
            h^{(j)}_{k^{\prime},k^{\prime}} & h^{(j)}_{k^{\prime},k^{\prime\prime}}\\            h^{(j)}_{k^{\prime\prime},k^{\prime}} & h^{(j)}_{k^{\prime\prime},k^{\prime\prime}}
        \end{bmatrix}
        \begin{bmatrix}
            t_j\alpha_j \\
            t_j(1-\alpha_j)
        \end{bmatrix}.
    \end{aligned}
\end{equation}

\begin{figure}
    \includegraphics[width=0.9\linewidth]{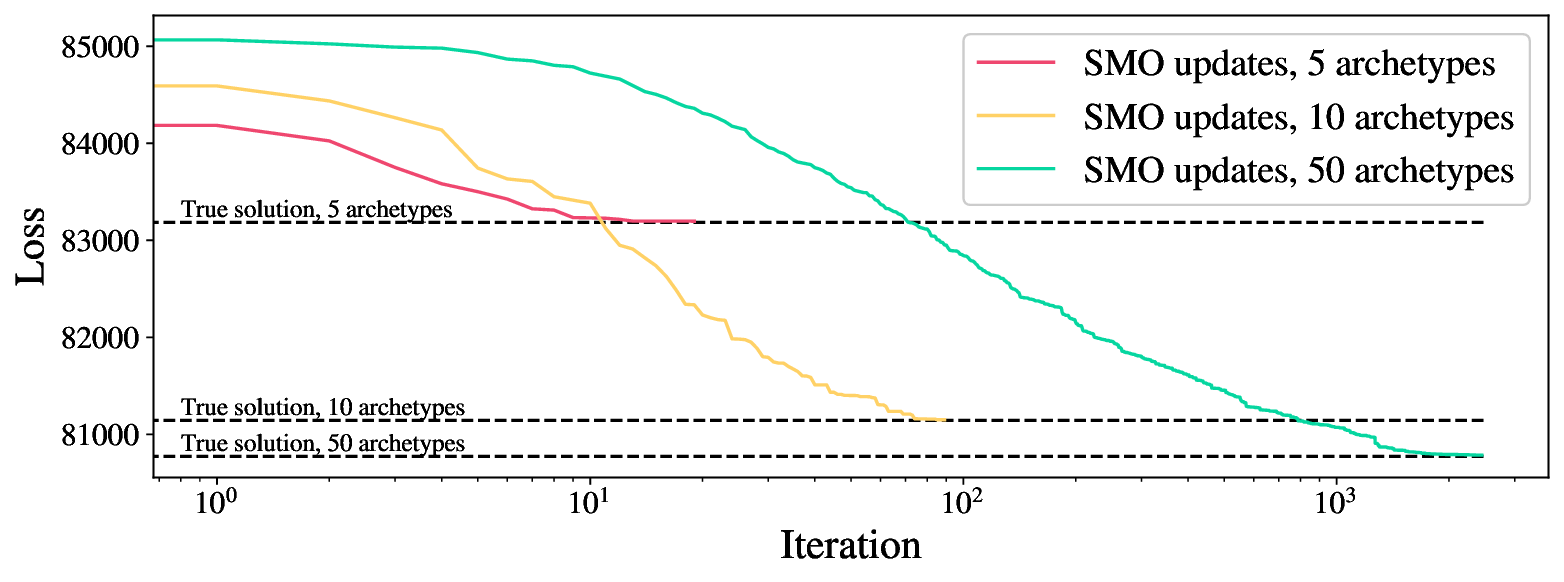}
    \caption{Comparison of SMO and FNNLS \cite{Bro1997AAlgorithm}. In all instances of K (number of archetypes), the SMO updates converges to the optimal solution within $K^2$ iterations. The optimal solution for the different $K$'s are marked by the dotted black lines identified by FNNLS.}
    \label{fig:SMO_convergence}
\end{figure}

This loss function is a quadratic function of  $\alpha_j$ which leads to the following closed-form update: 
\begin{equation} \label{eq:alpha}
    \alpha^{*}_j = -\frac{t_j h^{(j)}_{k^{\prime},2}+ \sum t_jh^{(j)}_{k^{\prime\prime},k^{\prime}}-2 t_j h^{(j)}_{k^{\prime\prime},k^{\prime\prime}}-2d_{k^{\prime},j}+2 d_{k^{\prime\prime},j}}{2t_j (h^{(j)}_{k^{\prime},k^{\prime}}-h^{(j)}_{k^{\prime},k^{\prime\prime}}-h^{(j)}_{k^{\prime\prime},k^{\prime}}+h^{(j)}_{k^{\prime\prime},k^{\prime\prime}})}.
\end{equation}
Since $\alpha^{*}_j$ is restricted to be between 0 and 1 we set $\alpha^{*}_j = 1$ if $\alpha^{*}_j >1$ and $\alpha^{*}_j = 0$ if $\alpha^{*}_j < 0$ which is optimal given the convexity of the loss function wrt. $\alpha_j$.
$\alpha_j^{*}$ thereby redistributes the weight of the archetypes between the two considered components. To demonstrate that this method works in an AA framework, the same function has been minimized using a QP solver. From figure \ref{fig:SMO_convergence} it is clear that the SMO updates converges to the correct solution after $K^2$ iterations corresponding in orders of magnitudes to covering all pairwise combinations of$K$ components and is thus highly efficient in the typical AA scenario in which few components are extracted from the data. 

\subsection{$\mathbf{C}$-update by an Active Set Procedure}

To optimize $\mathbf{C}$ a fast non-negative least squares algorithm (FNNLS)\cite{Bro1997AAlgorithm} is used. This enables us to utilize the sparsity of $\mathbf{C}$. As originally proposed by \cite{Bell2007ModelingSystems} we apply a linear constraint directly into an active set algorithm imposing a quadratic penalty $\lambda(1-\sum_j c_{jk})^2$. We include a small regularization $\epsilon\sum_j c_{jk}^2$ to ensure the associated Hessian of the QP is full rank. To define a suitable QP problem for the optimization method a Taylor expansion is applied to the likelihood, similar to the $\mathbf{S}$-update the general form becomes:

\begin{equation} \label{eq:Fc}
    f(\mathbf{c}_k) \approx const.-(\mathbf{d}_k+\lambda_k\mathbf{1})^\top\mathbf{c}_k+\frac{1}{2}\mathbf{c}_k^\top(\mathbf{H}^{(k)}+\lambda_k\mathbf{1}\mathbf{1}^\top +\epsilon_k\boldsymbol{I})\mathbf{c}_k
\end{equation}
in which $\lambda_k=10^{9}\cdot mean((\mathbf{H}_{A,A}^{(k)})^2)$, where $A$ is the active set and $\epsilon_k=\lambda_k\cdot 10^{-15}$. From the identified active set $A$ the solution is given in closed form as $\mathbf{c}_{A,k}=(\mathbf{H}^{(k)}+\lambda\mathbf{1}\mathbf{1}^\top +\epsilon_k\boldsymbol{I})_{A,A}^{-1}(\mathbf{d}_k+\lambda_k\mathbf{1})_A$ whereas entries not part of the active set is set to zero \cite{Bro1997AAlgorithm}. $\mathbf{d}$ and $\mathbf{H}$ for the specific distributions can be seen in equations eq. (\ref{eq:AS_d_LS})-(\ref{eq:AS_H_BER}). Notably, the active set procedure has cubic scaling in terms of the size of the active set. However, the $\mathbf{C}$ matrix is in general expected to be sparse, resulting in a small active set and an efficient procedure.
For least squares we obtain:
\begin{equation}\label{eq:AS_d_LS}
\begin{aligned}
         d_{j,k} &= \sum\limits_{i,j^\prime} x_{i,j^\prime} x_{i,j} s_{k,j}-x_{i,j^{\prime}}r_{i,j}s_{k,j} + \sum\limits_{j^{\prime},j^{\prime\prime}} h^{(k)}_{j^{\prime},j^{\prime\prime}}c_{j,k}, 
\end{aligned}
\end{equation}

\begin{equation}\label{eq:AS_H_LS}
    h^{(k)}_{j^{\prime},j^{\prime\prime}} = \sum\limits_{i,j} x_{i,j^\prime} x_{i,j^{\prime\prime}} s_{k,j}^2.
\end{equation}

and for the Bernoulli likelihood the expression becomes: 


\begin{equation}\label{eq:AS_d_BER}
    d_{j,k} = \sum\limits_{i,j^\prime}  p_{i,j}( \frac{x_{i,j^\prime}}{r_{i,j^\prime}} - \frac{(1-x_{i,j^\prime})}{(1-r_{i,j^\prime})})s_{k,j^\prime}+\sum\limits_{j^{\prime},j^{\prime\prime}} h^{(k)}_{j^{\prime},j^{\prime\prime}}c_{j,k},
\end{equation}

\begin{equation}\label{eq:AS_H_BER}
    h^{(k)}_{j_{\prime},j_{\prime\prime}} = \sum\limits_{i,j} p_{i,j^\prime}p_{i,j^{\prime\prime}}(\frac{x_{i,j}}{r_{i,j}^2} + \frac{(1-x_{i,j})}{(1-r_{i,j})^2})s_{k,j}^2.
\end{equation}

\subsection{Principal Convex Hull Algorithm for Bernoulli Likelihood}
For completeness, we propose a simple modification to the principal convex hull algorithm (PCHA) \cite{Mrup2012ArchetypalMining}, where we replace the least squares loss and derived gradients with the following gradients and loss based on the Bernoulli log-likelihood (eq. \ref{eq:Berloss}) denoted B-PCHA
\begin{equation}
    g_{j, k}^{C} = \sum\limits_{i,j^\prime}  p_{i,j}( \frac{x_{i,j^\prime}}{r_{i,j^\prime}} - \frac{(1-x_{i,j^\prime})}{(1-r_{i,j^\prime})})s_{k,j^\prime} ,
\end{equation}

\begin{equation}
    g_{k, j}^{S}  = \sum\limits_{i,j^\prime}  p_{i,j^\prime}( \frac{x_{i,j}}{r_{i,j}} - \frac{(1-x_{i,j})}{(1-r_{i,j})})c_{j^\prime,k}.
\end{equation}
Based on these gradients, the update rules become
\begin{equation}
c_{j, k} \leftarrow \max \left\{\tilde{c}_{j, k}-\mu_{\widetilde{C}}\left(g_{j, k}^{C} -\sum_{j^{\prime}} g_{j^{\prime}, k}^C \tilde{c}_{j^{\prime}, k}\right), 0\right\},
\end{equation}
\begin{equation}
s_{k, j} \leftarrow \max \left\{\tilde{s}_{k, j}-\mu_{\widetilde{S}}\left(g_{k, j}^{S}-\sum_{k^{\prime}} g_{k^{\prime}, j}^{S} \tilde{s}_{k^{\prime}, j}\right), 0\right\},
\end{equation}
where $\mu_{\widetilde{C}}$ and $\mu_{\widetilde{S}}$ are step-size parameters tuned by linesearch. $\mathbf{\widetilde{C}}$ and $\mathbf{\widetilde{S}}$ are the $\mathbf{C}$ and $\mathbf{S}$ matrices where the column stochastic constrains have been enforced by renormalizing the columns of  $\mathbf{C}$ and $\mathbf{S}$ to the standard simplex, see also \cite{Mrup2012ArchetypalMining} for details. 

\subsection{Normalized Mutual Information}
AA is in general unique \cite{Mrup2012ArchetypalMining} and to evaluate model consistency Normalized Mutual Information (NMI) is used. This method was introduced to AA in \cite{Hinrich2016ArchetypalData}. This method evaluates the columns in $\mathbf{S}$ as probability distributions and compares the consistency across 10 runs. The Mutual Information is normalized between 0 and 1, therefore an NMI of one indicates that the model converges to the same solution (up to permutation of the components) regardless of how the $\mathbf{C}$ and $\mathbf{S}$ matrices are initialized.

\section{Results and discussion}
\begin{figure*}[!htb]
\minipage{0.36\textwidth}
      \subfloat[Convergence\label{fig:realConvergence}]{%
          \includegraphics[width=\textwidth]{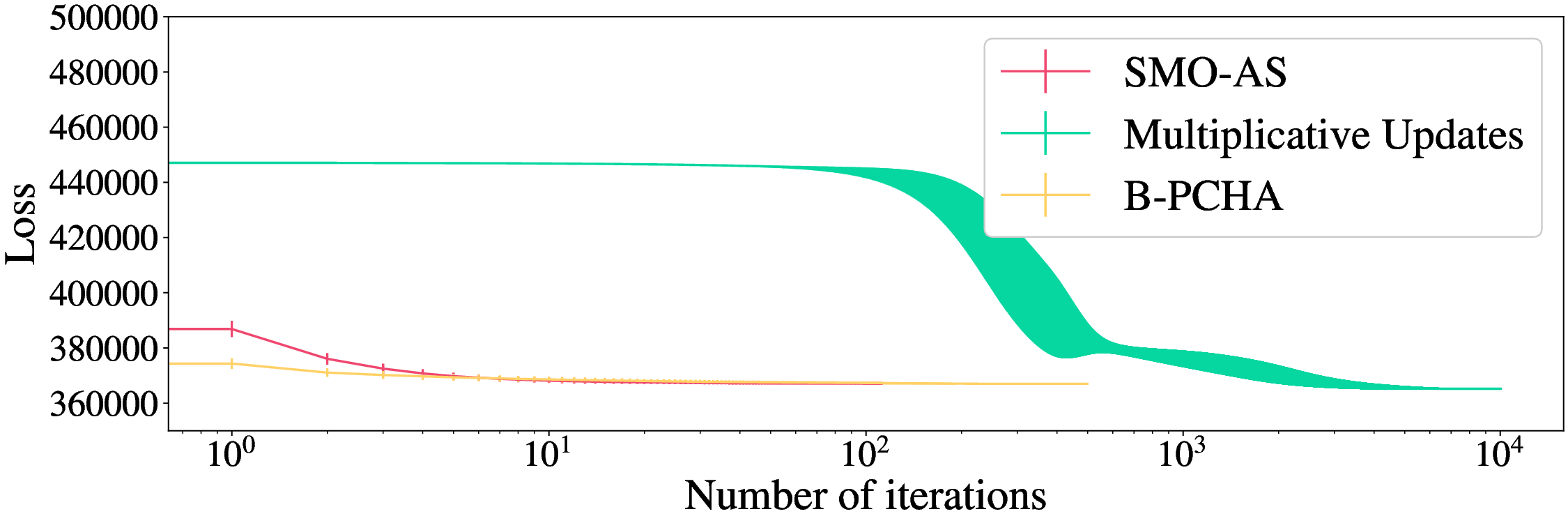}}\\
    \subfloat[Loss\label{fig:realLoss}]{%
          \includegraphics[width=\textwidth]{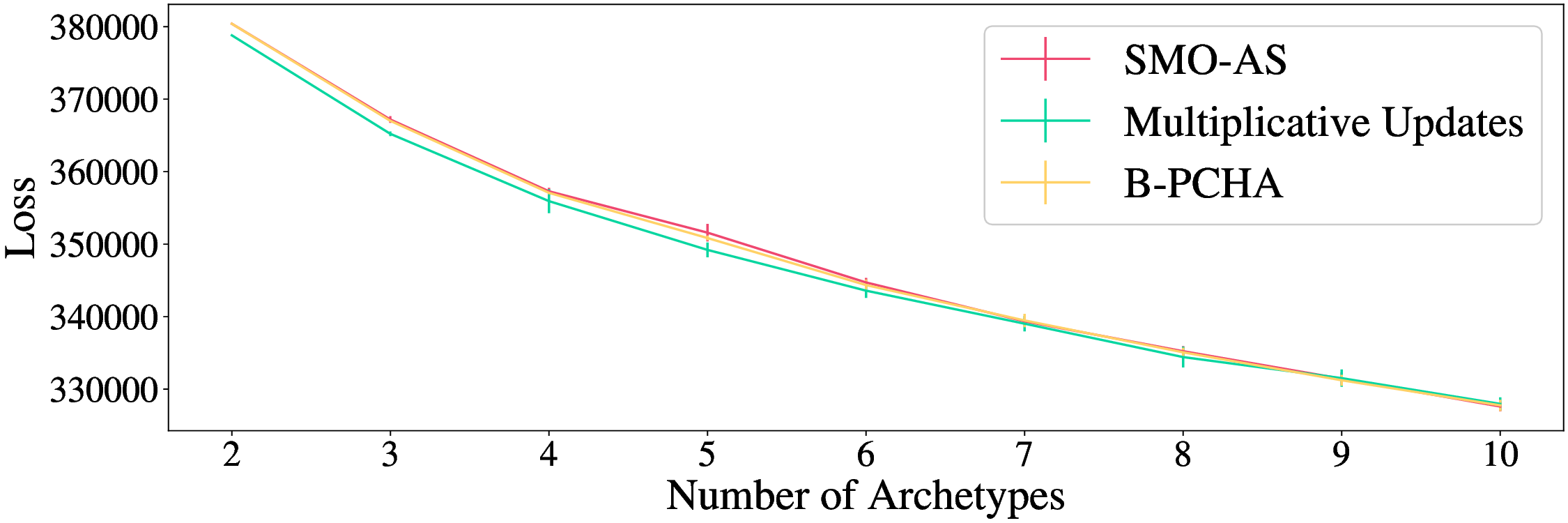}}
\endminipage
\minipage{0.36\textwidth}
      \subfloat[NMI\label{fig:realNMI}]{%
          \includegraphics[width=\textwidth]{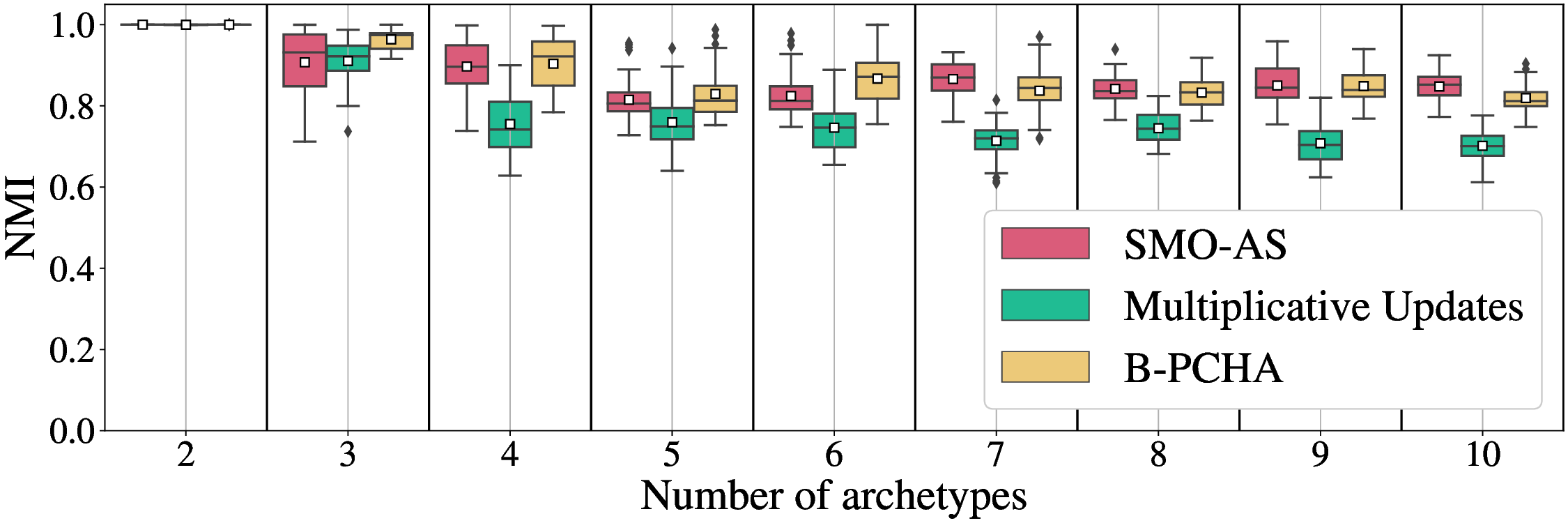}}\\
    \subfloat[Runtime\label{fig:realruntime}]{%
          \includegraphics[width=\textwidth]{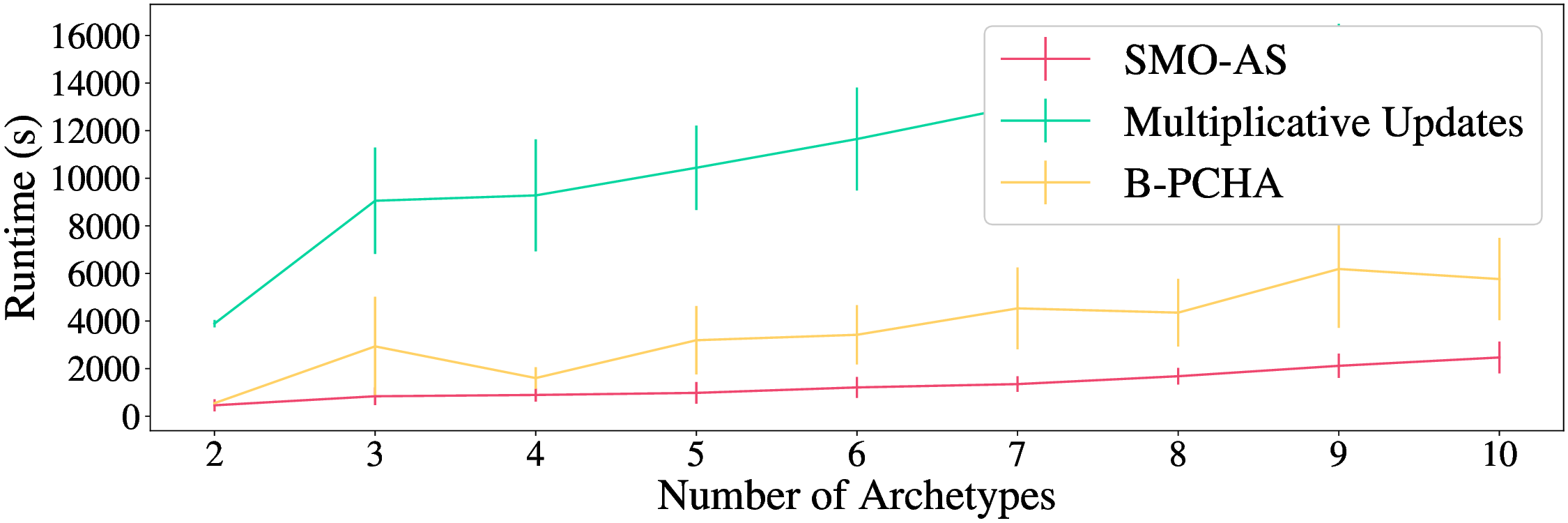}}
\endminipage
\minipage{0.265\textwidth}%
\subfloat[Simplex representation\label{fig:realsimplex}]{%
          \includegraphics[width=\textwidth]{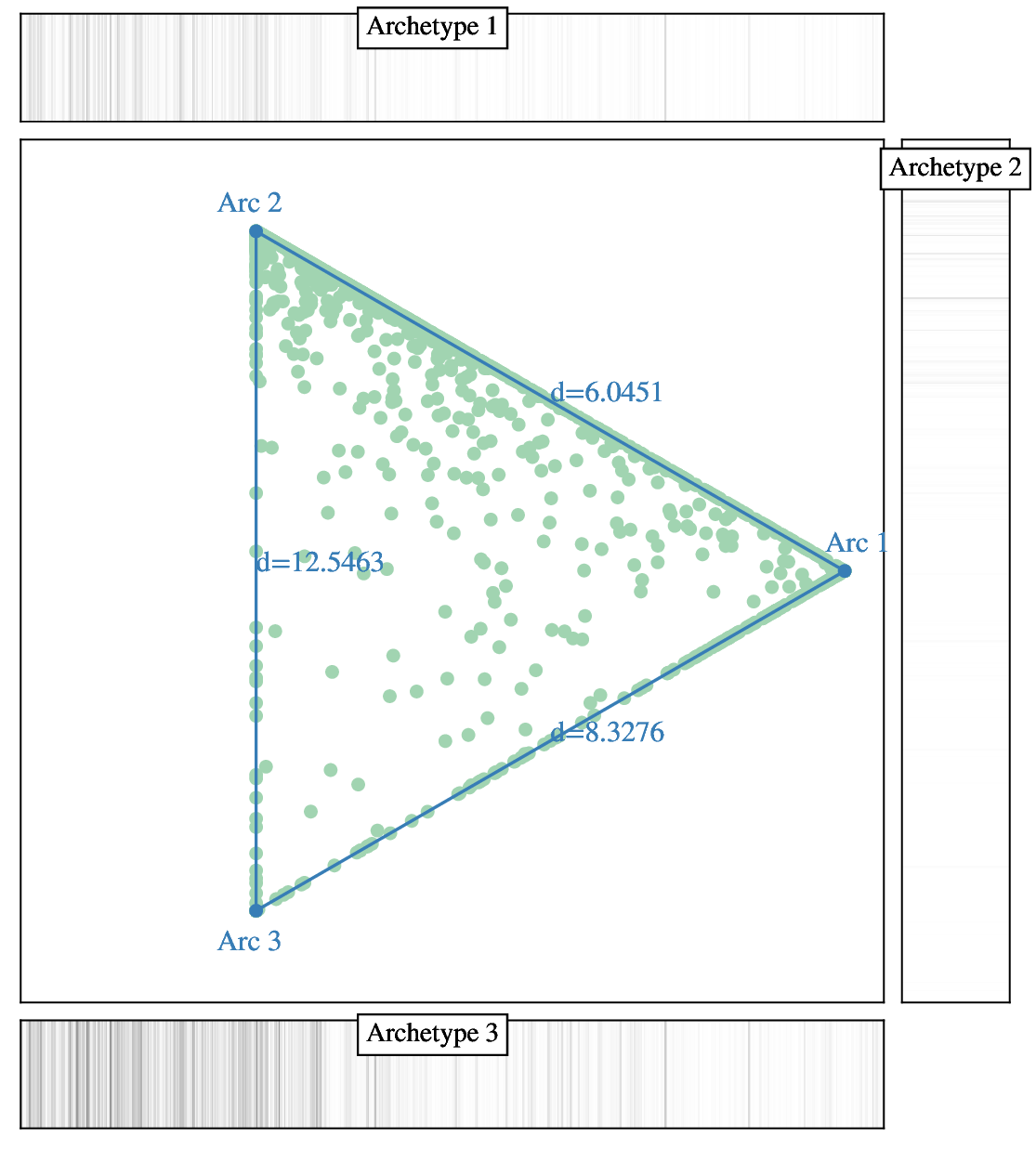}}
\endminipage
\caption{Performance visualizations on the SIDER data set. From (a) and (b) we observe that while SMO-AS and B-PCHA converges within fewer iterations, all three models converge to approximately the same loss and with comparable stability (NMI), although the solutions found by Multiplicative updates are less stable, especially for higher number of archetypes (c). From (d) it can be seen that the SMO-AS model clearly outperforms the other models in terms of runtime.}
\vspace{-3mm}
\label{fig:real}
\end{figure*}
To demonstrate the proposed optimization framework's efficiency and convergence properties we first evaluate it on two synthetic data sets respectively based on continuous data (Gaussian distribution likelihood corresponding to conventional least squares optimization) and binary data (Bernoulli distribution likelihood corresponding to cross-entropy minimization). The data was generated to have eight archetypes, 800 features and 1000 samples. We compare our proposed inference procedure to two existing AA inference methodologies - the PCHA algorithm for least squares (Gaussian) \cite{Mrup2012ArchetypalMining} and multiplicative update procedure for binary (Bernoulli) \cite{Seth2016ProbabilisticAnalysis} data. The results from the synthetic study can be seen in Fig.~\ref{fig:synth}. It is clear that our SMO-AS and B-PCHA frameworks exhibits faster convergence than the multiplicative updates \cite{Seth2016ProbabilisticAnalysis} while producing similar quality of loss-solutions ($L(\mathbf{X},\mathbf{R})$)  Fig.~\ref{fig:synth}(b) and (e) as well as model consistencies (NMI) Fig.~\ref{fig:synth}(c) and (f). We further evaluated the models on a binary data set of drugs and their side effects (SIDER) \cite{Kuhn2016TheEffects} in which each entry defines the presence or absence of a given side-effect for a given drug. The data matrix, after being filtered for the MedDRA preferred terms, contains 1347 samples (drugs) and 5868 features (side effects) with a sparsity of $98.3\%$. The results for the real dataset can be seen in Fig.~\ref{fig:real}. We again observe that our generic optimization procedures clearly outperforms the multiplicative update method in terms of speed and convergence. 

Determining the optimal number of archetypes remains an open problem.  For the simulated data in Fig. \ref{fig:synth} in which we know the true number of simulated archetypes we observe that the correct number of archetypes follows a regime in which the model substantially improves in loss ($L(\mathbf{X},\mathbf{R})$) by inclusion of archetypes whereas model consistency (NMI) is high. For the real dataset the number of archetypes considering these two aspects is less evident but inspecting $L(\mathbf{X},\mathbf{R})$ and NMI we observe that the three component models exhibits high NMI and substantial loss improvements. We therefore display the inferred model structure for $K=3$ components (Fig.~\ref{fig:real}(a) and (e)). 
In Fig~\ref{fig:real}(e) we observe that the archetypes primarily delineate drugs' tendencies to include side effects such that archetype 1 and 3 represent drugs with many as opposed to archetype 1 representing drugs with few side effects. Whether or not further insights about the archetypes can be made we leave for future work.

The code was run on a Intel Xeon Gold 6226R\cite{DTUComputingCenter2024DTUResources}. While both the B-PCHA and the SMO-AS models are shown to converge faster both in terms of speed and iterations (fig.~\ref{fig:real} (a) and (d)) when compared to the multiplicative updates it is nearly impossible to do a fair runtime comparison. Our SMO-AS framework allows for trivial parallelization and supports GPU acceleration. We have chosen not to include this in order to make the comparison as fair as possible and we find that the SMO-AS approach in run-time outperforms both the multiplicative updates and B-PCHA procedures. Notably, the Sequential Minimal Optimization (SMO) framework for $\mathbf{S}$ is remarkably efficient for a small number of archetypes. However, the efficiency of the active set algorithm used to update $\mathbf{C}$ heavily depends on the size of the active set and in situations where the active set becomes large our active set procedure is inefficient. One approach to control the maximal size of the active set could be to explore the furthestsum method proposed in \cite{Mrup2012ArchetypalMining} to select a reduced set of observations used to form the archetypes. This method can be used to substantially restrict the number of observations used to define the archetypes and thereby the maximal size of the active set, but it may render the model susceptible to archetypes simply defined by outliers.
Alternatively, the active set size could be restricted during the analysis such that the active set is only allowed to grow to a certain size. Future work should investigate how this can be implemented while ensuring convergence. The most important advantage of the proposed optimization frameworks for AA is their generality, enabling the optimization to work for arbitrary data distribution. This both has the benefit that the model is easy to use, but also that convergence is guaranteed from the convexity of the alternating $\mathbf{S}$ and $\mathbf{C}$ optimization problems and associated closed-form updates not relying on any hyper-parameter tuning such as gradient step sizes
\cite{Mrup2012ArchetypalMining} or necessitating likelihood tailored convergence derivations as required for multiplicative updates \cite{Seth2016ProbabilisticAnalysis, Lee2000AlgorithmsFactorization}. 
\section{Conclusion}
We presented two novel frameworks for archetypal analysis, one that leverages the structure of the data distribution to derive efficient closed-form updates for the AA model exploring that $\mathbf{C}$ when sparse can be efficiently inferred using an active set procedure whereas $\mathbf{S}$ when low-dimensional can be efficiently inferred using sequential minimal optimization (SMO). The second method expanded the PCHA framework to Bernoulli data. We showed how to instantiate our frameworks for continuous (Gaussian) and binary (Bernoulli) data and demonstrated its effectiveness on both synthetic and real. The SMO approach due to its $K^2$ scaling is especially suitable in the typical scenario where $K\leq 50$. The active set procedure to estimate C is suitable, provided that the active set remains small. In situations where this is not the case, we recommend using the proposed gradient-based B-PCHA approach for binary data. We also compared our approach with two prominent existing methods for archetypal analysis for Gaussian \cite{Mrup2012ArchetypalMining} and Bernoulli \cite{Seth2016ProbabilisticAnalysis} likelihood inference and confirmed that our procedure exhibits efficient optimization as function of alternating iterations updating $\mathbf{C}$ and $\mathbf{S}$. Our framework can be easily extended to other data distributions by use of appropriate likelihood functions.

\bibliographystyle{IEEEbib}
\bibliography{strings,refs,references}

\end{document}